\documentclass[10pt,twocolumn,letterpaper]{article}

\usepackage[pagenumbers]{cvpr} 

\usepackage{graphicx}
\usepackage{amsmath}
\usepackage{amssymb}
\usepackage{booktabs}
\usepackage{multirow}
\usepackage{makecell}
\usepackage{booktabs}
\usepackage{colortbl}
\usepackage{xcolor}

\usepackage{float}

\usepackage[pagebackref,breaklinks,colorlinks]{hyperref}

\usepackage[capitalize]{cleveref}
\crefname{section}{Sec.}{Secs.}
\Crefname{section}{Section}{Sections}
\Crefname{table}{Table}{Tables}
\crefname{table}{Tab.}{Tabs.}

\begin{document}

\title{Task Integration Distillation for Object Detectors}

\author{Hai Su$^{1}$\quad Zhenwen Jian$^{1}$\quad Songsen Yu$^{*}$\thanks{Corresponding author.}\\
South China Normal University\\
{\tt\small 2022024197@m.scnu.edu.cn}
}

\maketitle

\begin{abstract}
  Knowledge distillation is a widely adopted technique for model lightening. However, the performance of most knowledge distillation methods in the domain of object detection is not satisfactory. Typically, knowledge distillation approaches consider only the classification task among the two sub-tasks of an object detector, largely overlooking the regression task. This oversight leads to a partial understanding of the object detector's comprehensive task, resulting in skewed estimations and potentially adverse effects. Therefore, we propose a knowledge distillation method that addresses both the classification and regression tasks, incorporating a task significance strategy. By evaluating the importance of features based on the output of the detector's two sub-tasks, our approach ensures a balanced consideration of both classification and regression tasks in object detection. Drawing inspiration from real-world teaching processes and the definition of learning condition, we introduce a method that focuses on both key and weak areas. By assessing the value of features for knowledge distillation based on their importance differences, we accurately capture the current model's learning situation. This method effectively prevents the issue of biased predictions about the model's learning reality caused by an incomplete utilization of the detector's outputs.
  \par Extensive experiments demonstrate the significant potential and versatility of the proposed method. Notably, when employing Task Integration Distillation (TID) for feature decoupling, the pure distillation effect alone yields a stable improvement of approximately 2.0\% in mean Average Precision (mAP) for object detectors, surpassing recent feature decoupling methodologies and other feature distillation components in feature decoupling efficacy.
  
\end{abstract}

\section{Introduction}
\label{sec:intro}
With the advancement of deep learning technologies, an increasing array of models based on deep learning have been proposed and widely adopted across various fields. To achieve superior performance, deep learning models often resort to larger and more complex backbone networks, necessitating greater training resources and higher hardware specifications. To address this challenge, the technique of knowledge distillation was introduced \cite{hinton2015distilling}. Knowledge distillation is a method that transfers knowledge from a sizable teacher model to a lightweight student model without incurring additional cost, enabling the student model to achieve or even surpass the teacher model's performance. However, the majority of knowledge distillation approaches \cite{heo2019comprehensive, tung2019similarity,yim2017gift, zagoruyko2016paying} have been designed for image classification tasks, resulting in suboptimal effectiveness of knowledge distillation techniques within the domain of object detection.

As one of the fundamental tasks in the field of computer vision, object detection has increasingly garnered attention. However, object detection models, which require performing both regression and classification of targets, embody a relatively complex and challenging task, typically necessitating large and sophisticated network architectures. These models incur significant computational and storage costs, thus impeding their widespread deployment across various application scenarios. To align with the practical needs for lightweight object detection models, knowledge distillation techniques applied to these models have attracted considerable interest and development from the scholarly community.

Knowledge distillation in object detection mainly uses feature-based techniques \cite{wang2019distilling,guo2021distilling,chen2017learning, yang2022focal} to boost student model performance with features from intermediate layers.It is widely acknowledged that direct distillation from feature maps yields limited improvements. Thus, selecting features from feature maps that can effectively enhance the distillation effect is crucial. Feature-based knowledge distillation methods in object detection predominantly select features based on the ratio of image information noise, identifying key regions with a higher information ratio through techniques such as cropping, grading, and weighting functions. The concept of knowledge distillation originates from the real-world teaching process between teachers and students, where a teacher must not only select essential knowledge for students but also pay attention to the students' specific learning condition. Within the context of exploring the application of knowledge distillation in the field of object detection, this paper introduces an appropriate expression: Within the framework of knowledge distillation, we define the learning condition of the student model as its performance in terms of localization accuracy and classification precision of targets within images during the training process. However, the methods mentioned above focus primarily on the selection of key information without considering the detailed learning conditions of the student. Some researchers have supplemented this by utilizing the model's classification outcomes to refine the selection of key features, extracting non-target objects close to targets within the background areas through classification outcomes. This approach of selecting features based on classification outputs \cite{zhixing2021distilling}, \cite{li2022knowledge} considers the student's specific learning conditions to some extent.

However, object detection consists of two main subtasks: regression and classification, both critical for detector performance. Focusing solely on classification outcomes without considering regression results can lead to a skewed prediction of the student model's actual learning state. The skewed prediction refers to the issue where current knowledge distillation methods, unable to accurately capture the student model's learning state, may lead to a divergence in feature focus from the model's established tasks. This indicates that knowledge distillation approaches require further optimization to ensure they more accurately reflect the student model's learning effectiveness for object detection tasks. While some researchers have incorporated both classification and regression in knowledge distillation \cite{tang2023task}
, their aim was to address the discordance between classification and regression outputs in detection, not to discuss the model's learning condition.

It should be noted that knowledge distillation methods were initially developed within the realm of image classification before being adapted to object detection. However, there has been no discussion on the simultaneous application of classification and regression tasks to predict the student model's actual learning situation. Therefore, devising a feature knowledge distillation approach that encompasses both classification and regression tasks, aimed at reflecting the student model’s learning condition, represents an innovative challenge from the ground up.

To comprehensively consider the actual condition of the student model, we propose a feature knowledge distillation approach based on classification and regression tasks. To our knowledge, this is the first feature knowledge distillation method in the domain of object detection that addresses both classification and regression tasks, aimed at reflecting the model's real learning situation. We have defined a feature score to quantitatively assess each feature's performance in relation to classification and regression tasks. Drawing on the method used by object detectors to differentiate between positive and negative samples for regression boxes, we characterize the performance of teacher and student models in the regression task within knowledge distillation. Utilizing the feature scores, we identify key knowledge and the actual learning situation of the student model by defining areas of strength and weakness. Ultimately, we decouple the feature maps into high-value, medium-value, and low-value regions for separate processing. Additionally, we take into account the real learning situation derived from the outputs of the teacher and student models, which aids in further segmentation of features based on the learning condition of the student detector.

In summary, our contributions are as follows: 

\begin{itemize}
  \item
  We propose a novel approach to prevent skewed predictions of the student model's learning condition during the knowledge distillation process. This method simultaneously addresses classification and regression tasks, enabling accurate assessments of the student model's actual learning condition.
  \item
  We introduce a method to balance the selection of key knowledge with the specific learning condition of the student model. This strategy leverages detector outputs from areas of strength and weakness, selecting critical features for improved learning focus.
  \item
  We validate the effectiveness and versatility of our method through extensive experiments on the MSCOCO and VOC datasets with various backbone networks and across different object detectors, outperforming other knowledge distillation approaches that utilize model outputs. 
\end{itemize}

\begin{figure*}
  \centering
  \includegraphics[width=0.97\linewidth]{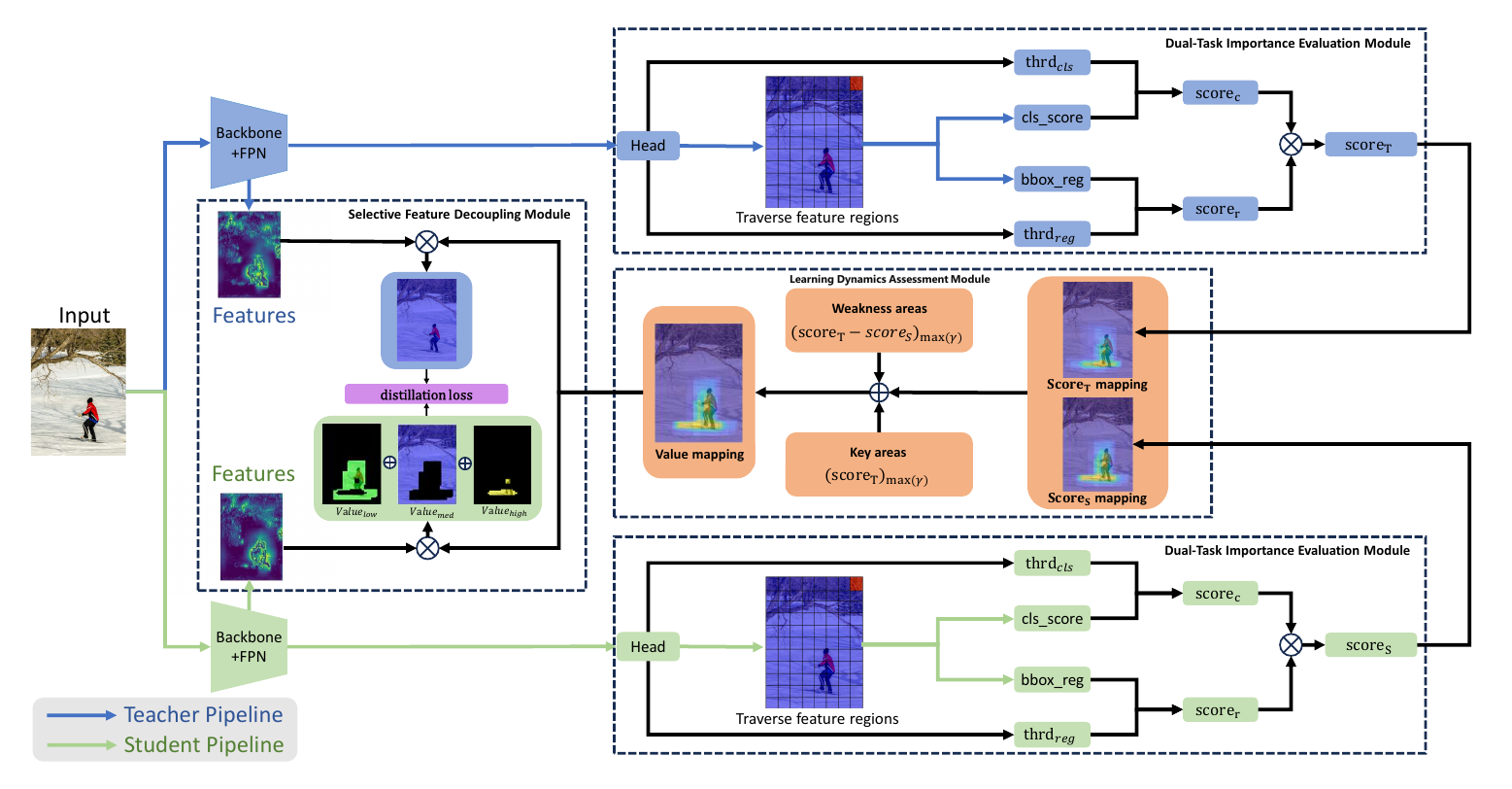}
  \caption{Details the Task Integration Distillation (TID) approach, which includes the Dual-Task Importance Evaluation Module for quantifying model output results, the Learning Dynamics Assessment Module that reflects the model's current learning condition based on output value, and the Selective Feature Decoupling Module that decouples feature maps according to the learning condition. For concise representation of the methodological flow, only single-level features and predictions in the FPN are shown here.}
  \label{figure:structure}
\end{figure*}

\section{Related Work}
\label{sec:related work}

\subsection{Knowledge distillation}

Knowledge distillation is an effective model compression technique that facilitates the transfer of knowledge from a large model to a smaller one, enabling the smaller model to achieve, or even approximate, the performance of the larger model. Initially introduced by Hinton et al. \cite{hinton2015distilling}, knowledge distillation has been predominantly applied in the image classification domain. The technique primarily transfers "dark knowledge" to the student model through the soft labels of the teacher model. To smoothly extract this "dark knowledge”, a hyperparameter known as temperature is introduced.

Knowledge distillation was initially proposed in the domain of image classification and has since been widely applied across various fields. Currently, knowledge distillation can be categorized into three types based on the source of knowledge: response-based knowledge distillation \cite{cho2019efficacy, furlanello2018born, mirzadeh2020improved, yang2019snapshot, zhang2018deep}, relation-based knowledge distillation \cite{park2019relational, peng2019correlation}, and feature-based knowledge distillation \cite{heo2019comprehensive, heo2019knowledge, huang2017like, kim2018paraphrasing}. Response-based knowledge distillation, the earliest proposed method, involves extracting the output of the teacher network's last layer and directly mimicking the teacher's final prediction. Relation-based knowledge distillation extracts relationships between different layers or data samples within the teacher network, transferring such relational knowledge to the student to achieve distillation. Feature-based knowledge distillation involves extracting and aligning feature maps from both the teacher and student models to the same size, then having the student model's feature maps fit those of the teacher model, facilitating the transfer of knowledge. Due to its ease of handling knowledge sources and almost uniform performance across most domains, feature-based knowledge distillation has garnered widespread attention from scholars for its strong versatility.

Knowledge distillation methods are not only widely applied in the field of computer vision, such as image retrieval \cite{wu2022contextual} and face recognition \cite{li2023rethinking}, but also in natural language processing \cite{tang2019distilling, liu2020cross}, speech recognition \cite{kunzmann2021exploiting}, recommendation systems \cite{tan2016improved}, information security \cite{zhi2021lightweight}, multimodal data \cite{zhang2022unims}, and finance and securities \cite{hayashi2019long}.
 Knowledge distillation, with its capability to transfer knowledge between different models, has garnered significant attention across multiple domains due to its versatility.

\subsection{Object detection}
Object detection is one of the core tasks in computer vision, primarily aimed at locating and classifying objects within images. With the advancement of Convolutional Neural Networks (CNNs), current object detection methods can mainly be divided into three categories: two-stage detectors \cite{cai2018cascade, girshick2015fast, ren2015faster, he2017mask}, anchor-based one-stage detectors \cite{lin2017focal, liu2016ssd, redmon2018yolov3}, and anchor-free one-stage detectors \cite{tian2019fcos, yang2019reppoints}.

One-stage detectors can directly obtain object classifications and predicted bounding boxes on feature maps, consuming fewer resources compared to two-stage detectors. Two-stage detectors typically first generate region proposals using a Region Proposal Network (RPN), followed by classification and refinement of bounding boxes in the second stage. Although this approach often yields better results, it correspondingly increases resource consumption.

Anchor-based detectors can directly predict object classes and bounding boxes from feature maps, making them more efficient than two-stage detectors. However, anchor-based detectors still utilize a large number of predefined anchor boxes as reference points, increasing the computational cost of the detector. To reduce this computational expense, anchor-free one-stage detectors were introduced. Anchor-free one-stage detectors can directly predict objects' keypoints and locations without using anchor boxes, but their performance is generally not as high as that of anchor-based detectors.

Despite significant differences in model structures and improvement strategies among modern object detectors like YOLO \cite{redmon2016you}, GFL \cite{li2020generalized}, and ATSS \cite{zhang2020bridging}, all require processing images through a backbone network to obtain features, which are then processed in the object detection head. While different detectors have different detection heads, their inputs are features extracted by the backbone network. Our feature-based knowledge distillation approach acts solely on features and is only related to the classification and regression tasks of object detection, making our method broadly applicable and not limited by changes in detectors. Therefore, our knowledge distillation method, designed around the two subtasks of object detection, exhibits good versatility and can work across different detectors. Knowledge distillation allows for the substitution of simple backbone networks for complex ones without sacrificing model performance and can be used alongside other model lightening methods. Typically, the size of a model is directly proportional to the costs associated with deploying and using the model. Lightweight models can be deployed in low-configuration settings to save costs, making further lightening of current object detectors of significant value.

\subsection{Knowledge Distillation for Object Detection}
Chen et al. were among the first to apply knowledge distillation to the field of object detection \cite{chen2017learning}. Initially, it processed the model's outputs as soft labels to facilitate the transfer of "dark knowledge" between the teacher and student models, thereby enhancing the student model's performance. Existing knowledge distillation methods can generally be categorized into response-based, feature-based, and relation-based approaches. Among these, feature-based knowledge distillation methods, due to their stronger versatility, have garnered widespread attention in the domain of object detection.

In the domain of object detection, FitNet \cite{hinton2015distilling} demonstrated that feature information from intermediate layers also contributes to guiding the student model. FGFI \cite{wang2019distilling} progressively improved the use of intermediate layer features to enhance the performance of student detectors.

In the process of knowledge distillation for detectors, traditional methods have attempted to select appropriate distillation areas from the perspective of knowledge distillation. Initially, FitNet used all intermediate layer features for knowledge distillation \cite{hinton2015distilling}, but due to the imbalance between target and background areas in object detection, extracting all features introduced a significant amount of noise, leading to poor distillation effects. To mitigate the imbalance between targets and backgrounds, feature selection began to be implemented. FGFI \cite{wang2019distilling} selected target areas using true bounding boxes, but this method discarded background areas containing a lot of noise and partial information after selecting target areas, failing to fully utilize all feature regions. Defeat \cite{guo2021distilling} conducted distillation separately for background and target areas without discarding the background, further proving the utility of background information, i.e., that optimal performance can be achieved through appropriate feature selection methods. Recently, FGD \cite{yang2022focal}[33] introduced a global distillation module and further decoupled features through spatial and channel attention. MGD \cite{yang2022masked} attempted to combine feature imitation with feature generation, providing a new perspective for knowledge distillation.

Furthermore, there have been attempts to utilize the output results of detectors for knowledge distillation. GID \cite{dai2021general} proposed distillation areas in an instance-wise manner through classification results, while PFI \cite{li2022knowledge} and FRS \cite{zhixing2021distilling} introduced prediction-guided distillation using classification scores. TFD \cite{tang2023task} applied teacher model output and localization results to generate masks, addressing the issue of inconsistent task distribution. However, these methods tend to focus more on partial outputs of detectors, and the incomplete use of output results can lead to a tendency that contradicts the objective of the task, such as emphasizing the classification task while neglecting the localization task. Additionally, these methods struggle to reflect the model's actual learning condition, and the incomplete utilization of output results can lead to skewed predictions about the model's actual learning situation by knowledge distillation methods.

The primary distinction between our proposed Task Integration Distillation (TID) and the aforementioned works lies in our approach to the current issue. We introduce a feature selection method that accommodates both classification and regression tasks. Drawing inspiration from real-world teaching, where teachers not only select key knowledge for student learning but also pay attention to the students' specific learning conditions, we propose a distillation approach focusing on areas of strength and weakness that fully reflects the model's actual learning condition, thereby mitigating the issue of skewed predictions in the model.

\section{Method}
\label{sec:method}

In previous studies, most knowledge distillation methods did not fully consider the model's classification and regression outputs. This partial treatment of model outputs could lead to biased predictions regarding the model's learning condition through knowledge distillation. To address this issue, we introduce a generic module named Task Integration Distillation (TID). This section will systematically present the overall architecture of TID.

As illustrated in \cref{figure:structure}, our proposed Task Integration Distillation (TID) comprises three main components: the Dual-Task Importance Evaluation Module (DIEM), the Learning Dynamics Assessment Module (LDAM), and the Selective Feature Decoupling Module (SFDM). The DIEM is responsible for extracting the model's output results and quantifying these results to assess the output value of each feature point. Notably, the output value is determined by both the model's classification and regression outputs, ensuring consideration of both subtasks and avoiding skewed predictions about the model's learning condition. The LDAM accurately assesses the model's current learning condition. Drawing on strategies from the real-world teaching process, it uses the output values from both the teacher and student models to delineate areas of strength and weakness, thereby more accurately reflecting the model's current learning state. Finally, the SFDM evaluates the importance of features based on the model's learning condition and decouples features into high-value, medium-value, and low-value features according to the learning condition of the student model.

In summary, our proposed Task Integration Distillation (TID) framework ascertains the significance of each feature by holistically analyzing the model's classification and regression outputs. By categorizing areas based on task significance—identifying regions of strength and weakness—it reflects the model's learning condition. This comprehensive approach to knowledge distillation is aimed at augmenting the model's performance and efficiency. Detailed discussions on each sub-module will be provided in Sections 3.1, 3.2, and 3.3, respectively.

\subsection{Dual-Task Importance Evaluation Module}

In this section, we will progressively detail the derivation process of the Task Importance Evaluation Module. As previously mentioned, our aim in selecting key areas for knowledge distillation is to consider both the classification and regression tasks of the detector. Thus, it is essential first to quantify the output results of each feature point, which we refer to as the task importance of the feature point. Typically, the extraction of task importance encompasses two aspects: quantifying the detector's localization results and quantifying the detector's classification results. As shown in Figure 3, we divide the feature map processed by the object detection model into regions based on feature points, with each region obtaining corresponding classification and regression results. To ensure that knowledge distillation takes into account both subtasks and avoids skewed predictions about the model's learning condition, we reprocess the output results using the criteria for maximum suppression of classification results, thrd\_cls, and the division of positive and negative samples for bounding boxes, thrd\_pos, set by the target detection model. This results in corresponding task scores, and finally, the sum of these task scores is used as a reference for the importance of the feature region in the model score.

\begin{figure}
  \centering
  \includegraphics[width=0.9\linewidth]{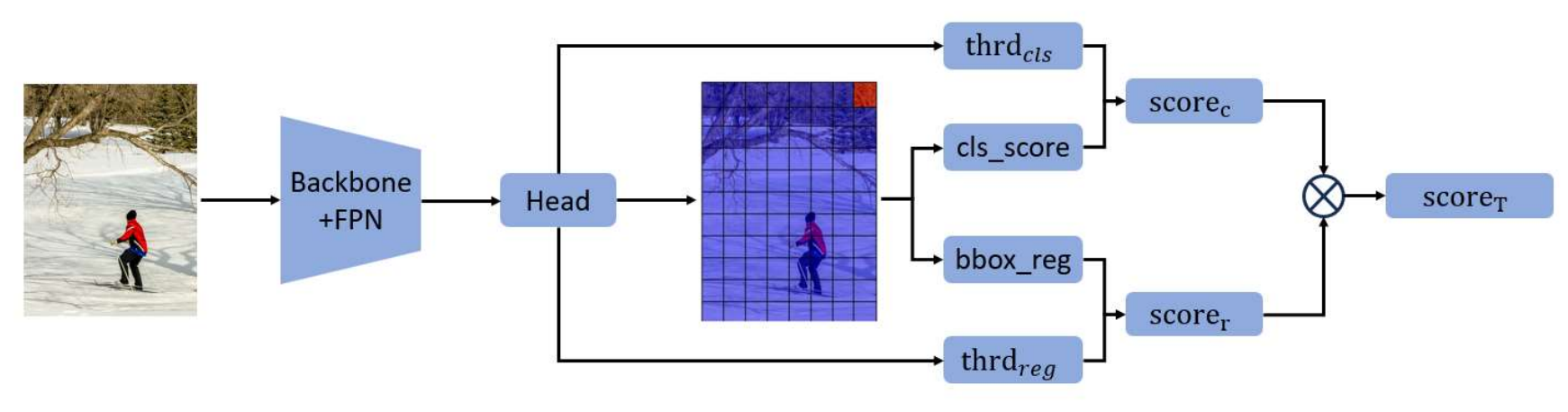}
  \caption{ details the Dual-Task Importance Evaluation Module. For clarity, we display operations on a single feature map. The red regions represent the currently processed sub-regions of the feature map, with each sub-region receiving an area importance score based on classification and regression outputs, as well as the intrinsic settings of the object detection model.}
  \label{figure:heat map}
\end{figure}

For the detector's localization function, we quantify the detector's localization results with $Score_r$. Initially, we obtain the maximum Intersection over Union($IOU_{\max}$) for the bounding boxes generated by each feature point. Subsequently, based on the division relationship between positive and negative samples in mainstream detectors, we divide the feature points into three distinct intervals according to $IOU_{\max}$, and assign the corresponding $Score_r$, with the specific formula as follows:

\begin{equation}
    Score_r = 
    \begin{cases} 
    2, & \text{if}\ IOU_{\max} \geq thrd\_pos \\
    1.5, & \text{if}\ thrd\_neg < IOU_{\max} < thrd\_pos \\
    0.4, & \text{if}\ IOU_{\max} \leq thrd\_neg
    \end{cases}
\end{equation}
Where $IOU_{\max}$ signifies the highest Intersection over Union between the bounding boxes generated by feature points and the ground truth boxes. $thrd\_pos$ and $thrd\_neg$  represent the positive and negative sample thresholds established by the detector for classifying samples. In alignment with the detector's definitions of positive and negative samples, we assign higher localization scores to feature points with an $IOU_{\max}$ exceeding the positive sample threshold, and lower localization scores to those with an $IOU_{\max}$ below the negative sample threshold. The specific thresholds for positive and negative samples are determined based on the detector's model.

For the detector's classification functionality, we quantify the detector's classification results with $Score_r$. To circumvent limitations imposed by specific detector architectures, we simply retrieve the classification output for each feature point and compare it with the classification result of the current ground truth box for the corresponding category. Given the varying difficulty of classification across different images and the variation in image sizes, employing a fixed threshold for $Score_r$ could result in an uneven distribution of feature points across images. Consequently, we designate the top 2.5\% of the classification scores as positive classification samples. The remainder are considered negative classification samples, as expressed by the following formula:
The value of \(Score_c\) is determined as follows:
\begin{equation}
    Score_c = 
    \begin{cases} 
    1.5, & \text{if}\ (i,j) \in \text{thrd\_cls} \\
    1, & \text{otherwise}
    \end{cases}
\end{equation}
Where $thrd\_cls$ indicates that the classification score of a feature point is within the top 2.5\% of all feature points, $otherwise$ signifies that the classification score of the feature point does not fall within this top 2.5\%.

At the conclusion of this module, we integrate the classification and localization scores to derive a score that represents the importance of a feature point with respect to both subtasks. The formula is as follows:
\begin{equation}
    Score = Score_c \times Score_r
\end{equation}
Where \(Score_c\) and \(Score_r\) respectively denote the classification score and the localization score, \(Score_T\) and \(Score_S\) represent the importance scores of the teacher model and the student model, respectively.

Diverging from previous studies, we not only assess the output value of feature points from the perspective of knowledge distillation but also consider the detector's classification and localization output methods. It's important to note that, currently, we have only obtained the detector's classification and localization scores. In the following section, we will explore how to reflect the model's specific learning conditions through classification scores and select key areas for knowledge distillation based on both classification and localization scores. This approach will aid in gaining a more comprehensive understanding of the model's learning condition and provide more effective guidance for knowledge distillation.

\subsection{Learning Dynamics Assessment Module}
Within the framework of knowledge distillation, we define the learning condition of the student model as its performance in localizing and accurately classifying targets within images during the training process. Drawing inspiration from real-world teaching scenarios and the intrinsic properties of detector outputs, we propose a feature selection method. This method balances key and weak areas, accurately capturing the student model's learning condition, and selects areas crucial for knowledge distillation.

As previously mentioned, this paper maps the relationship between teachers, students, and examinations in real life to the process of knowledge distillation. We observe that in actual teaching processes, teachers not only need to impart key knowledge to students but also need to pay attention to students' learning progress and comprehension. Inspired by this, we attempt to apply this teaching model to the knowledge distillation process of detectors. This approach is expected to enhance the learning efficiency and effectiveness of models.

This paper introduces a feature selection module based on key and weak areas. In the teacher model, the task importance of feature points is determined based on the output and classification results of the teacher model. Feature points with higher task importance are considered to be of greater significance to the detector's output results. Drawing from the real-world analogy of the relationship between teachers and exams, the information contained within these feature points is deemed by the teacher model as valuable for enhancing the detector's performance, thus necessitating focused attention from the student model on these points. We have selected feature points that the teacher model considers to contain more information, and we define the feature areas composed of these points as key areas. The specific formula is expressed as follows:
\begin{equation}
    Score_{\text{key}} = \left( Score_{T} \right)_{\max(\gamma)}
\end{equation}
Where \(Score_T\) signifies the classification and localization scores for feature points within the teacher model, and \(Score_{\text{key}}\) indicates the key score of the feature points. Consequently, areas with elevated key scores are deemed more significant, and conversely for areas with lower scores.

The weak areas defined in this paper refer to regions composed of feature points that possess high task importance in the teacher model but are deemed less important in the student model. Typically, feature-based knowledge distillation necessitates the transfer of information from the feature maps of the teacher model to the student model. Our proposed measure of task importance enables both teacher and student models to evaluate the contribution of a feature point to the detector's output. Due to differences in model performance, feature points assessed by the teacher model as highly contributory often contain more information, whereas the student model's evaluation might not be as precise. Regions where the student model's assessment significantly diverges from that of the teacher model usually indicate inadequacies in the student model's extraction of information from these parts, preventing it from aligning with the teacher model's judgment. We refer to these regions as weak areas. The specific formula is expressed as follows:
\begin{equation}
    Rrea_{\text{weak}} = \left( Score_{(i,j)}^{\text{teacher}} - Score_{(i,j)}^{\text{student}} \right)_{\max(\gamma)}
\end{equation}
Where \(Rrea_{\text{weak}}\) represents the weak areas, \(Score_{(i,j)}^{\text{teacher}}\) indicates the classification and localization scores on the corresponding feature map in the teacher model; \(Score_{(i,j)}^{\text{student}}\) denotes the classification and localization scores on the corresponding feature map in the student model. \(i\) and \(j\) represent the horizontal and vertical coordinates, respectively. \(\max(\gamma)\) signifies the selection of the top \(\gamma\) regions with the largest difference.

We quantify the importance of key and weak areas to better manage the relationship between these two regions. Key areas are represented by key scores, thus we also quantify the weak areas. The specific formula is expressed as follows:
\begin{equation}
Score_{\text{weak}} = 
\begin{cases} 
1.5, & \text{if } (i,j) \in Rrea_{\text{weak}} \\
1, & \text{otherwise}
\end{cases}
\end{equation}
Where \(Rrea_{\text{weak}}\) indicates that a feature point is within a weak area, and Otherwise signifies that a feature point does not reside within a weak area, \(Rrea_{\text{weak}}\) quantifies the weak areas. 

It is noteworthy that we do not suppress features outside the weak areas, as the concept of weak areas is specifically introduced for the student model. Our preference is to direct the student model's attention towards features within weak areas, without implying that features in other areas should not be learned.

Furthermore, due to the performance disparity between the teacher and student models, we observe that feature points with significant differences are often concentrated around hard and small samples. This is attributed to the student model's challenges in accurately localizing and classifying small and difficult samples. Therefore, we employ Non-Maximum Suppression (NMS) to ensure that weak areas are not solely focused on hard and small samples.

Ultimately, after identifying key and weak areas through the detector's output results, we amalgamate these two areas to calculate the value of each feature point. Based on this value, we can determine whether a feature point falls within the critical areas that require knowledge distillation. This constitutes a detector output-oriented method for selecting areas. The specific formula for calculating value is as follows:
\begin{equation}
    Value_{\text{output}} = Score_{\text{key}} \times Score_{\text{weak}}
\end{equation}
Where \(Rrea_{\text{key}}\) and \(Rrea_{\text{weak}}\) represent the scores for key and weak areas respectively, \(Value_{\text{output}}\) denotes the output value of the feature point. Based on the value of the feature points, we select the key areas for distillation.

\subsection{Selective Feature Decoupling Module}

In this section, we will detail how we select pivotal areas through the Selective Feature Decoupling Module. The feature distillation approach captures information from various levels of the backbone network post-feature fusion via the FPN, utilizing diverse feature selection methods and segmentation strategies to facilitate the transfer of knowledge from the teacher model to the student model. Typically, the current prevalent methods for feature decoupling spatially separate features based on ground truth (GT), which can be represented as follows:
\begin{align}
    L_{\text{fea}} ={}& \alpha_{\text{bg}} M_{\text{bg}} \left( F_{(i,j)}^T - f\left(F_{(i,j)}^S\right) \right) \nonumber \\
    &+ \alpha_{\text{obj}} M_{\text{obj}} \left( F_{(i,j)}^T - f\left(F_{(i,j)}^S\right) \right)
\end{align}
Where  \(F^T\) and \(F^S\) denote the feature maps of the teacher model and the student model, respectively, \(\alpha_{\text{bg}}\) and \(\alpha_{\text{obj}}\) represent the hyperparameters utilized for balancing the contributions of background and object features. The function f is employed to adjust \(F^S\) to match the dimensions and scale of \(F^T\). \(M_{\text{bg}}\) and \(M_{\text{obj}}\) are masks applied for selective feature extraction.

Prior research \cite{guo2021distilling} has demonstrated that decoupling features into multiple components based on differences among them can effectively enhance the outcomes of knowledge distillation. Compared to background areas, target areas typically contain more information and less noise. Thus, building on these findings, we have developed a novel approach to knowledge distillation that decouples based on the output value and informational value of feature points.

Anchors near target areas contain a greater wealth of usable information, while background areas are more prone to higher noise levels. Hence, we adjust the output value to progressively decrease from target areas towards background areas. The informational value is delineated as follows: 

\begin{equation}
V_{\text{information}} = 
\begin{cases} 
1 & \text{if } (i,j) \in \partial \\
\frac{\sqrt{(a_i - b_i)^2}}{\text{dists}} & \text{if } (i,j) \in \beta \\
0 & \text{otherwise}
\end{cases}
\end{equation}
Where \(V_{\text{information}}\) denotes the informational value, \(i\) and \(j\) are the horizontal and vertical coordinates on the feature map, respectively. \(\alpha\) represents the area within the ground truth bounding box, and \(\beta\) signifies the vicinity around the ground truth box. a and b correspond to the coordinates of the anchor and the closest ground truth box, respectively. \(dists\) refers to the maximum distance within the area surrounding the ground truth box.

Upon determining the informational value, we combine the output value derived from key and weak areas with the informational value to ascertain the value of the feature points, specifically expressed as follows:
\begin{equation}
V_{ij} = V_{\text{output}} \times V_{\text{information}}
\end{equation}
Where \(V_{\text{output}}\) represents the output value of the feature points obtained from key and weak areas, \(V_{\text{information}}\) denotes the informational value of the feature points acquired through traditional decoupling methods, and i,j signify the horizontal and vertical coordinates on the feature map. \(V_{ij}\)  embodies the final value derived for each feature.

Previous research has indicated that finer granularity in feature decoupling can significantly enhance the performance of knowledge distillation. Following this premise, we decouple the feature map into three segments based on the value hierarchy, namely, high-value areas, medium-value areas, and low-value areas. The formula is detailed as follows:
\begin{equation}
M_{ij} = 
\begin{cases} 
v_{ij}, & \text{if } v_{ij} \geq 1 \\
0.7 \times v_{ij}, & \text{if } 0.01 \leq v_{ij} < 1 \\
0.05 \times v_{ij}, & \text{if } v_{ij} < 0.01
\end{cases}
\end{equation}
Where \(V_{ij}\) represents the value of the feature points, we categorize the feature points into three sections based on their value. The value of feature points in low-value areas is suppressed, whereas the value of feature points in high-value areas remains unaffected.

Ultimately, we establish our feature decoupling method, which segments the features into three areas based on their value hierarchy, expressed in the formula as: 
\begin{align}
L_{\text{fea}} ={}& M_{\text{high}} \left( F_{(i,j)}^T - f\left(F_{(i,j)}^S\right) \right) \nonumber \\
&+ M_{\text{med}} \left( F_{(i,j)}^T - f\left(F_{(i,j)}^S\right) \right) \nonumber \\
&+ M_{\text{small}} \left( F_{(i,j)}^T - f\left(F_{(i,j)}^S\right) \right)
\end{align}
Where \(M_{\text{high}}\) represents the mask for high-value areas, \(M_{\text{med}}\) for medium-value areas, and \(M_{\text{small}}\) for low-value areas. \(\alpha_{\text{high}}\), \(\alpha_{\text{med}}\), and \(\alpha_{\text{low}}\) denote the balancing coefficients for these three areas, respectively.

\section{Experiments}
\label{sec:experiments}
\subsection{Dataset}
To validate the universality of our proposed knowledge distillation method, we conducted experiments on two common object detection datasets: MS COCO 2017 and the VOC dataset. Specifically, our experiments were primarily conducted on MS COCO 2017, with the VOC dataset serving to demonstrate the generality of our approach. We employed average precision (mAP) and average recall (mAR) as evaluation metrics, including \(mAP\), \(mAR\), \(AP_{50}\), \(AP_{75}\), \(AR_{50}\), \(AR_{75}\), \(AP_{S}\), \(AP_{M}\), and \(AP_{L}\). Where \(AP\) primarily indicates the error rate in prediction results, and \(AR\) reflects the rate of missed detections within those predictions. \(AR_{S}\), \(AR_{M}\), and \(AR_{L}\) are utilized to evaluate performance across objects of varying scales, specifically targeting small, medium, and large objects, respectively, to gauge the detector's efficacy. \(AP\) primarily reflects the error rate of prediction results, and \(AR\) indicates the rate of missed detections in the predictions. The thresholds of 50 and 75 denote that detection boxes are considered positive when the IOU exceeds 50 and 75, respectively.

We conducted experiments across various object detection frameworks, including both anchor-based and anchor-free models, to demonstrate the universality of our feature selection method. Our methodology was compared with other knowledge distillation approaches within the GFL \cite{li2020generalized} and ATSS \cite{zhang2020bridging} object detection frameworks. All experiments were implemented using MMdetection and PyTorch. Our configurations adhere to the default settings in MMdetection, with a learning rate of 0.02, utilizing SGD with a momentum set to 0.09, batch size to 4, and weight decay to 0.9. All experiments followed a 1x training schedule, equivalent to 12 epochs. Furthermore, all student models were pretrained on ResNet, and unless specified otherwise, we did not employ any inheritance strategies. Our experiments were carried out on an A30 GPU.

\subsection{Main results}
In this section, we compare our proposed Task Integration Distillation (TID) with other methods. The primary comparative experiments were conducted using two detectors, namely the GFL and ATSS object detectors. To demonstrate the superiority of our approach, we compared it with other methods that employ feature decoupling for knowledge distillation, as shown in \cref{table:main results}. 

\begin{table*}
  \centering
  \begin{tabular}{c|l|lccc|cc|lccc}
    \toprule
    Teacher& Student & mAP  & AP$_{S}$ & AP$_{M}$ &AP$_{L}$& AP$_{50}$ & AP$_{75}$ & mAR & AR$_{S}$ & AR$_{M}$ & AR$_{L}$\\
    \midrule
    \multirow{4}{*}{\makecell{Gfl\\ResNet101}}
    &Gfl-Res50 & 40.0 &21.8&43.6&52.0&57.8&43.0&57.8&35.8&62.6&74.1\\
    &Fitnet & 41.2(+1.2)&23.4&45.2&54.2&59.0&44.5&58.9(+1.1)&37.2&63.6&75.1\\
    &FGFI & 41.7(+1.7)&24.2&46.0&54.3&59.6&45.2&59.5(+1.7)&39.0&64.4&76.4\\
    &Defeat & 41.3(+1.3)&23.6&45.5&54.3&59.0&44.8&59.3(+1.5)&38.1&64.4&76.3\\
    &Gibox & 41.7(+1.7)&24.7&46.0&53.9&59.7&45.2&59.5(+1.7)&38.5&64.3&75.8\\
    &TFD & 41.7(+1.7)&24.4&45.9&54.8&59.7&45.1&59.6(+1.6)&39.1&64.1&76.2\\
    &LD & 41.8(+1.8)&23.7&45.9&54.7&59.7&45.0&59.4(+1.6)&38.2&64.6&76.1\\
    &\cellcolor{lightgray!45}Ours & \cellcolor{lightgray!45}42.0(+2.0)&\cellcolor{lightgray!45}24.8&\cellcolor{lightgray!45}46.3&\cellcolor{lightgray!45}54.8&\cellcolor{lightgray!45}59.9&\cellcolor{lightgray!45}45.7&\cellcolor{lightgray!45}59.7(+1.9)&\cellcolor{lightgray!45}39.7&\cellcolor{lightgray!45}64.6&\cellcolor{lightgray!45}76.6\\
    \midrule
    \multirow{4}{*}{\makecell{Gfl\\ResNet101}}
    &Gfl-Res18 & 28.2 &14.4&29.8&39.7&42.3&30.0&49.0&26.1&52.6&68.4\\
    &FGFI & 28.5(+0.3) &14.4&30.2&39.8&42.5&30.6&49.3(+0.3)&26.0&53.1&68.5\\
    &Defeat & 28.9(+0.7) &14.9&30.8&40.1&43.0&30.8&49.8(+0.8)&26.8&53.7&68.6\\
    &\cellcolor{lightgray!45}Ours & \cellcolor{lightgray!45}29.4(+1.2)&\cellcolor{lightgray!45}15.4&\cellcolor{lightgray!45}31.2&\cellcolor{lightgray!45}40.5&\cellcolor{lightgray!45}43.9&\cellcolor{lightgray!45}31.4&\cellcolor{lightgray!45}50.2(+1.2)&\cellcolor{lightgray!45}27.6&\cellcolor{lightgray!45}54.1&\cellcolor{lightgray!45}68.7\\
    \midrule
    \multirow{3}{*}{\makecell{Atss\\ResNet101}}
    &ATSS-Res50 & 40.5 &23.1&44.3&52.1&58.4&43.8&59.2&37.1&64.3&75.3\\
    &FGFI & 41.2(+0.7)&24.2&45.4&52.8&59.2&44.7&59.8&38.2&65.0&76.0\\
    &Gibox & 41.4(+0.9)&25.2&45.4&53.0&59.3&44.8&60.2&40.2&65.3&76.5\\
    &\cellcolor{lightgray!45}Ours  &\cellcolor{lightgray!45}41.5(+1.0)&\cellcolor{lightgray!45}24.3&\cellcolor{lightgray!45}45.4&\cellcolor{lightgray!45}53.5&\cellcolor{lightgray!45}59.4&\cellcolor{lightgray!45}44.8&\cellcolor{lightgray!45}60.2&\cellcolor{lightgray!45}38.6&\cellcolor{lightgray!45}64.9&\cellcolor{lightgray!45}77.0\\
    \midrule
  \end{tabular}
  
  \caption{presents the comparison results of knowledge distillation methods that utilize model output on the MS COCO 2017 dataset, between the proposed approach and existing methods utilizing model outputs.}
  \label{table:main results}
  
\end{table*}

As illustrated in \cref{table:main results}, through our feature decoupling method, we achieve superior performance compared to previous feature decoupling approaches. Under the guidance of the teacher model, our method demonstrates higher performance and better stability compared to other feature decoupling methods that rely solely on classification or localization results. Additionally, we conducted comparative experiments with other knowledge distillation methods that utilize model output results, as well as some feature-based knowledge distillation approaches, to validate the effectiveness of using model outputs to reflect learning conditions. Furthermore, we tested results under various teacher-student configurations to showcase the generalizability of our method across different backbone networks.

\begin{table}
  \centering
  \begin{tabular}{@{}c|ccc}
    \toprule
    Datasets& Distill & mAP  & AP$_{S}$ \\
    \midrule
    Pascal VOC    &- &0.7642 &0.764 \\
        &\checkmark &0.785 &0.7851 \\
    \midrule
    MS COCO    &- &0.4 &0.218 \\  
        &\checkmark &0.42 &0.248 \\
    \bottomrule
  \end{tabular}
  \caption{Results of Applying TID on Two Datasets.}
  \label{table:COCO_voc result}
\end{table}

In the field of object detection, in addition to the commonly used COCO dataset, we also applied our knowledge distillation method on another widely utilized dataset, VOC. This approach is predicated on the actual learning condition of the prediction model. By implementing our method on these two datasets, we were able to more thoroughly understand and evaluate the universality of our approach in practical applications, as shown in \cref{table:COCO_voc result}.

\subsection{Ablation study}
In this section, we conducted extensive experiments using the GFL object detector to demonstrate the effectiveness of each component and explore some implementation details of our method. Our analysis experiments were based on the teacher model ResNet-101 and the student model ResNet-50. All experiments were carried out with a training duration of 12 epochs and conducted on the MS COCO 2017 dataset.

\subsubsection{Ablation Study on Balancing Classification and Regression Tasks in the Dual-Task Importance Evaluation Module}

Our method is predicated on the classification and regression outcomes of the detector. Moreover, the assessment of the student model's learning condition necessitates consideration of both classification and regression tasks. Therefore, we conducted ablation experiments on the significance of classification and regression tasks for knowledge distillation, analyzing their importance to the distillation process. We considered knowledge distillation for classification and regression tasks separately while keeping the subsequent modules unchanged, as shown in \cref{table:task ablation study}.

\begin{table}
  \centering
  \begin{tabular}{@{}l|c|ccc}
    \toprule
    Method & \multicolumn{4}{c}{Gfl ResX101-Res50}\\
    \midrule
    L$_{cls}$  & - &\checkmark&-&\checkmark\\
    L$_{reg}$ & - &-&\checkmark&\checkmark\\
    \midrule
    mAP & 40.1 &40.9&40.9&{\bf42.0}\\
    AP$_{S}$ & 21.8 & 23.0&23.1&{\bf24.8}\\
    AP$_{M}$ & 43.6 &45.2&44.9&{\bf46.3}\\
    AP$_{L}$ & 52.0 & 53.8&53.6&{\bf54.8}\\
    \midrule
    mAR & 57.8 &58.7&58.7&{\bf59.7}\\
    AR$_{S}$ & 35.8 &36.9&36.7&{\bf39.7}\\
    AR$_{M}$ & 62.6 &63.4&63.6&{\bf64.6}\\
    AR$_{L}$ & 74.1 &75.4&75.8&{\bf76.6}\\
    \bottomrule
  \end{tabular}
  \caption{Ablation Study on Classification and Regression Tasks on the COCO Dataset.}
  \label{table:task ablation study}
\end{table}

We found that focusing exclusively on one type of output resulted in nearly identical performance for both tasks. This indicates that classification and regression tasks play almost equal roles in knowledge distillation. Furthermore, the effectiveness of estimating the model's learning condition significantly diminishes when relying on a single subtask as opposed to a combined assessment from both subtasks. This also highlights the issue of skewed predictions when the actual learning condition of the student model is determined based on just one task. Therefore, considering both classification and regression tasks to ascertain the student model's actual learning condition can mitigate the problem of biased predictions due to neglecting subtasks, thereby enhancing the effectiveness of knowledge distillation.

\subsubsection{Ablation Study on Integration of Regression Tasks within the Dual-Task Importance Evaluation Module}

In the context of effectively reflecting the model's learning condition by integrating both classification and regression tasks, it is crucial to further investigate how to balance the classification and regression tasks of an object detector. Initially, we discarded the approach of manually designing weights to determine the importance of the two tasks, as different types of object detectors handle classification and regression tasks in varied ways, making manually designed weights not universally applicable across multiple object detectors. Hence, we opt to use the evaluation criteria inherent to the object detector itself to determine the importance of the task in knowledge distillation. Specifically, we utilize the threshold for determining positive and negative samples in the detector's localization functionality as the source for the localization score. We believe that binding the localization score with the settings intrinsic to the object detector itself can enhance the universality of our method across different object detectors. Moreover, referencing the object detector's own assessment of the importance of different subtasks allows for a better integration of both subtasks.

Therefore, we designed comparative experiments to validate the impact on module performance of using scores derived directly from IOU versus those obtained through detector settings, as illustrated in \cref{table:IOU ablation}. These experiments compare the use of IOU directly for predicting learning conditions against using IOU scores obtained by combining IOU with the current object detector's positive and negative sample division for learning condition prediction. We observed that maintaining consistency with the detector's positive and negative sample division can yield greater benefits for knowledge distillation.

\begin{table*}
  \centering
  \begin{tabular}{@{}c|ccccccccccc}
    \toprule
    Method& IOU$_{score}$ & mAP  & AP$_{S}$ & AP$_{M}$ &AP$_{L}$& AP$_{50}$ & AP$_{75}$ & mAR & AR$_{S}$ & AR$_{M}$ & AR$_{L}$\\
    \midrule
    TDD*   &- &42.0 &24.8 &46.3 &54.8 &59.9 &45.7 &59.7 &39.7 &64.6 &76.6\\
    TDD*  &\checkmark &40.9 &23.5 &45.1 &53.1 &58.8 &44.2 &58.7 &37.4 &63.6 &75.1\\
    \bottomrule
  \end{tabular}
  \caption{Ablation Study on Using Detector's Positive and Negative Sample Division for IOU. Here, * indicates that the importance module for the localization task within the TID module has been removed.}
  \label{table:IOU ablation}
\end{table*}

We discovered that binding the localization score with the detector's settings can effectively enhance performance, indicating that the localization scores tied to the detector's configurations hold more value than those directly calculated using IOU. Given that knowledge distillation originated in the image classification domain, for the classification task, we will calculate the classification score based on the classification task's own output to reflect the importance of the classification task.

\subsubsection{Ablation Study for the Learning Dynamics Assessment Module}

Within the context of exploring the application of knowledge distillation in the field of object detection, this paper proposes an apt formulation: within the framework of knowledge distillation, we define the learning condition of the student model as its performance in terms of localization accuracy and classification accuracy on targets within images. We identify key and weak areas based on the task importance of different feature regions delineated by the object detector to reflect the model's learning condition. Through ablation studies, we validate the effectiveness of this segmentation method. As shown in \cref{table:key and weak ablation study}, when decoupling feature areas, considering either key areas or weak areas separately can lead to effective improvements, and combining both can further enhance the effects of knowledge distillation.

\begin{table}[H]
  \centering
  \begin{tabular}{@{}l|c|ccc}
    \toprule
    Method & \multicolumn{4}{c}{Gfl ResX101-Res50}\\
    \midrule
    L$_{weak}$  & - &\checkmark&-&\checkmark\\
    L$_{key}$ & - &-&\checkmark&\checkmark\\
    \midrule
    mAP & 40.1 &41.8&41.6&{\bf42.0}\\
    AP$_{S}$ & 21.8 & 24.5&24.2&{\bf24.8}\\
    AP$_{M}$ & 43.6 &46.2&45.6&{\bf46.3}\\
    AP$_{L}$ & 52.0 & 54.4&54.5&{\bf54.8}\\
    \midrule
    mAR & 57.8 &59.6&59.3&{\bf59.7}\\
    AR$_{S}$ & 35.8 &39.4&37.8&{\bf39.7}\\
    AR$_{M}$ & 62.6 &64.5&64.1&{\bf64.6}\\
    AR$_{L}$ & 74.1 &75.9&75.9&{\bf76.6}\\
    \bottomrule
  \end{tabular}
  \caption{Ablation Study for Key and Weak Areas, where \(L_{\text{weak}}\) and \(L_{\text{key}}\) represent the consideration of corresponding areas in calculating feature value.}
  \label{table:key and weak ablation study}
\end{table}

\subsection{Analysis and visualization}
\subsubsection{Error Type Analysis}
\begin{figure*}
  \centering
  \includegraphics[width=0.97\linewidth]{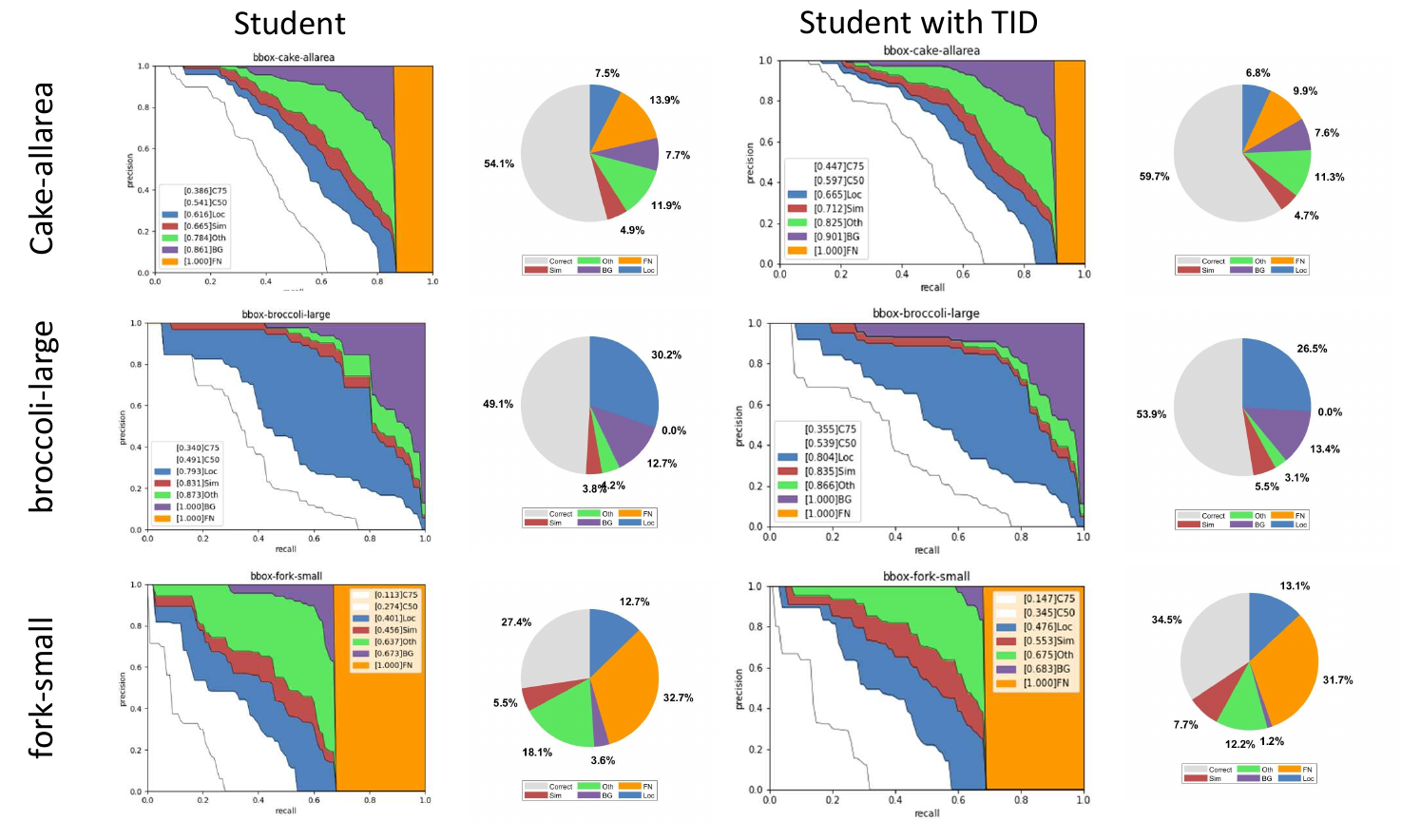}
  \caption{PR curves and error analysis between different models. 'Correct': Predictions with the correct label and an IOU greater than 0.5; 'Oth': False positives between classes, i.e., predictions with incorrect labels; 'FN': Missed detections; 'Sim': Predictions with incorrect labels but correct supercategories; 'BG': False alarms predicted in background areas; 'Loc': Predictions with the correct label but an IOU between 0.1 and 0.5.}
  \label{figure:err}
\end{figure*}

The COCO toolkit facilitates testing object detection models, providing outputs for model accuracy and proportions of different error types. We utilized the COCO toolkit to perform error analysis on object detectors before and after processing with our method. As illustrated in \cref{figure:err}, we report the accuracy of the two models and the proportions of various error categories. Following knowledge distillation using our method, the student model exhibits a significant reduction in localization errors, missed detections, and false positives between classes. This indicates our approach effectively focuses the knowledge distillation process on the student model's key and weak areas, enhancing the differentiation between background and targets, as well as between distinct classes. Utilizing the detector's output results further ameliorates the student model's localization issues. This confirms that TID enhances knowledge distillation in the manner we intended and validates the efficacy of TID.

\subsubsection{Focus Area Visualization}
In this section, we visualize the feature value proposed in our study within images.

\begin{figure*}
  \centering
  \includegraphics[width=0.97\linewidth]{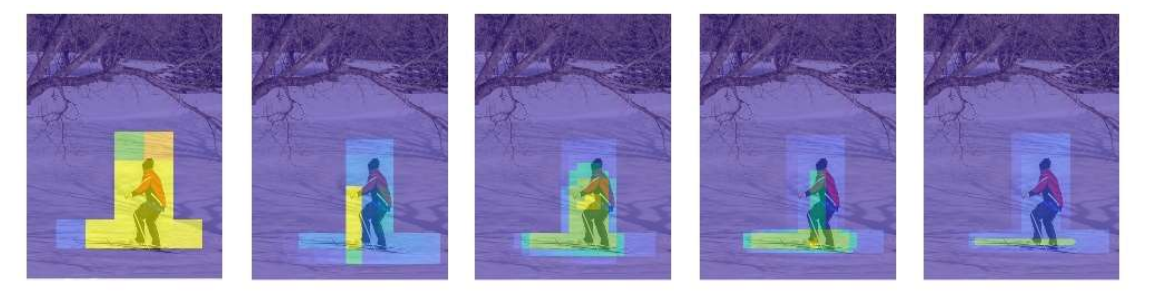}
  \caption{from left to right, displays the feature areas selected based on the model's learning condition across different FPN layers. The far left represents the lowest level features of the FPN, while the far right corresponds to the highest level features of the FPN.}
  \label{figure:visual}
\end{figure*}

As illustrated in \cref{figure:visual}, we visualized the specific value of each feature in different feature maps to better understand the importance of various features. We observed that the regions extracted by TID conspicuously cover most of the target areas. While focusing on target areas, TID pays additional attention to some edge regions and small target areas. This is because large targets are generally easier to recognize, whereas small targets and edge areas are more prone to errors. The TID we propose dynamically focuses on these areas, which are not only key but also represent weak points for the student model, based on the specific learning condition of the model.

\subsubsection{Qualitative Comparisons of Detector Outputs Before and After TID Processing}

\begin{figure*}
  \centering
  \includegraphics[width=0.97\linewidth]{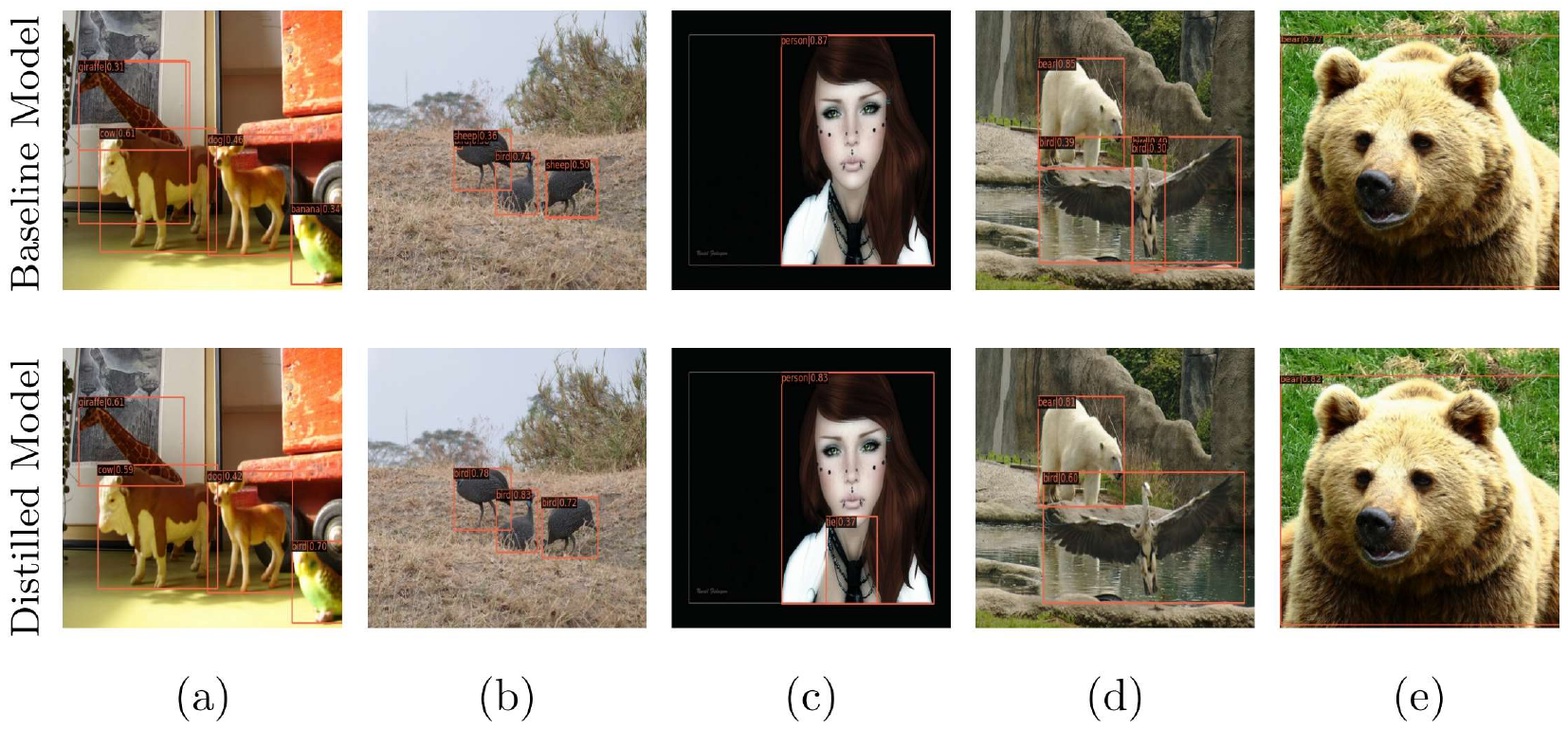}
  \caption{Qualitative analysis of the COCO2017 dataset using the baseline detector and the GFL-ResNet50 detector processed with our method.}
  \label{figure:detail err}
\end{figure*}

We conducted a qualitative comparison between the outputs of object detectors before and after processing with our method, as shown in \cref{figure:detail err}. The top row displays the results from the original object detectors, while the bottom row presents the outputs from the student detectors processed with our method. As illustrated in Figure 5(d), the student detector generated multiple detection boxes for the same target, whereas our method accurately recognized the entirety of the same target, avoiding the production of superfluous bounding boxes. Figure 5(a) shows the student model mistakenly identifying a bird as a banana, while our method correctly recognizes it as a bird. This demonstrates our method's ability to accurately pinpoint and address the weak points of the student detector. In Figure 5(b), the student detector incorrectly identified the background as a sheep, but our method avoided this error and produced a correct instance. In Figure 5(c), the student detector failed to detect a tie, whereas our method enabled the student detector to recognize the tie, indicating that the TID method enhances the student detector's discriminative abilities. Although both student detectors correctly identified the bear in Figure 5(e), the bounding box generated by the student detector using our method is closer to the bear, illustrating that our method can improve the quality of the bounding boxes produced by the student detector.

\section{Conclusion}
\label{sec:conclusion}
In this paper, we introduce a knowledge distillation method aimed at compressing object detection models. Our analysis highlights the existing knowledge distillation approaches' lack of utilization of classification and localization outputs, leading to an inability to accurately understand the current learning condition of the student model. Based on this insight, we propose the TID method to analyze learning conditions based on detector outputs and extract important features according to these conditions, thereby enhancing the performance of the student detector. The TID method is grounded in the FPN feature network, making it broadly applicable to object detection methods based on this network.

\nocite{*}

{\small
\bibliographystyle{ieee_fullname}
\bibliography{egbib}
}

\end{document}